\documentclass{article}

\PassOptionsToPackage{numbers, compress}{natbib}

     \usepackage[final]{neurips_2020}

\usepackage[utf8]{inputenc} 
\usepackage[T1]{fontenc}    
\usepackage{hyperref}       
\usepackage{url}            
\usepackage{booktabs}       
\usepackage{amsfonts}       
\usepackage{nicefrac}       
\usepackage{microtype}      
\usepackage{graphics}
\usepackage{amsmath}

\title{Spectral Unmixing With Multinomial Mixture Kernel and Wasserstein Generative Adversarial Loss}

\author{%
  Savas Ozkan, Gozde Bozdagi Akar\\
  Department of Electrical and Electronics Engineering\\
  Middle East Technical University\\
  Ankara, Turkey \\
}

\begin{document}

\maketitle

\vspace{-15pt}
\begin{abstract}
\vspace{-10pt}
This study proposes a novel framework for spectral unmixing by using 1D convolution kernels and spectral uncertainty. High-level representations are computed from data, and they are further modeled with the Multinomial Mixture Model to estimate fractions under severe spectral uncertainty. Furthermore, a new trainable uncertainty term based on a nonlinear neural network model is introduced in the reconstruction step. All uncertainty models are optimized by Wasserstein Generative Adversarial Network (WGAN) to improve stability and capture uncertainty. Experiments are performed on both real and synthetic datasets. The results validate that the proposed method obtains state-of-the-art performance, especially for the real datasets compared to the baselines. Project page at: \url{https://github.com/savasozkan/dscn}. 
\end{abstract}

\vspace{-5pt}
\section{Introduction}
\vspace{-10pt}
Spectral data provides comprehensive knowledge about the Earth's surface in the form of narrow-band spectra. The limitations of spectral sensors induce pure materials in a scene to be mixed in different fractions. Indeed, the foremost concern is that pure materials (i.e., endmembers) $\mathbf{E}=[\mathbf{e}_1,\mathbf{e}_2,...,\mathbf{e}_K] \in \mathbb{R}^{D \times K}$ and their fractions $\mathbf{y}= [y_1, y_2,...,y_K] \in \mathbb{R}^{K}$ have to be estimated blindly from spectral data $\mathbf{x} \in \mathbb{R}^{D}$. Furthermore, nonlinear relations and spectral uncertainty are ubiquitous with severe scattering conditions (see Appendix~\ref{appendix:su}).

Even if the mixture of data can be represented with linear/nonlinear formulations, these assumptions underestimate the solution by considering pure materials as a finite set towards spectral uncertainty. To overcome the limitations, the mathematical formulation is redefined as follows in~\cite{zhou2018gaussian}:
\begin{equation}
\mathbf{x} = \sum_{k=1}^{K} (\mathbf{e}_k + \mathbf{\gamma}_k) y_k + \eta = \mathbf{E}\mathbf{y} + \eta \text{, s.t. } y_k \geq 0, \sum_{k=1}^{K} y_k = 1 
\label{eq:sulinear}
\end{equation}

Here $K$ is the number of endmembers, and $D$ is the spectral bands of a pixel. Also, $\eta \sim \mathcal{N}(0,1)$  is the additive Gaussian noise to account for possible noise sources. Note that $\mathbf{\gamma}_{k}$ varies for each pixel and material. Hence, this term virtually defines the uncertainty of the data.

In the scope of this paper, we will present a novel framework that broadly uses this equation by computing fractions from pre-computed material signatures (i.e., only fractions are estimated from data). Specifically, we introduce a multinomial mixture kernel based on a neural network (NN) structure that obtains the fractions per-pixel. The learnable parameters of this kernel are optimized with WGAN for the first time. This step achieves theoretically more accurate and stable solutions.

\vspace{-5pt}
\section{Related Work}
\vspace{-10pt}
As expected, the initial methods are based on linear models due to their simplicity. In particular, sum-to-one and non-negative constraints are conditioned for least-square solutions~\cite{heylen2011fully}. However, these models are insufficient to resolve complex cases such as multiple scattering effects and microscopic-level material mixtures. Hence, bilinear models are used to handle nonlinear scenarios in which bilinear interactions of linear relations are considered in the models~\cite{halimi2011nonlinear, altmann2011bilinear}.

Even if the nonlinearity is retained, spectral uncertainty is still discarded in these models due to the finite set of materials. Distribution-based models represent endmember signatures as they are sampled from some distributions.  In the literature, various distributions are exploited~\cite{zhang2014pso, du2014spatial}. Notice that these models approximate each fraction to a unimodal distribution. Ultimately, this undermines the representation capacity as expected. Furthermore, the Gaussian Mixture  Model  (GMM) is adopted~\cite{zhou2018gaussian}. However, the complexity and assumptions limit the applicability of this method for large-scale data.

Recently, AutoEncoder (AE) network is redefined with additional layers and a spectral angle distance metric~\cite{ozkan2018endnet} to unmix data that improves robustness for spectral uncertainty. Moreover,~\cite{su2019daen} uses a stacked deeper AE network and a robust initialization is achieved by matrix factorization. 

\vspace{-5pt}
\section{Proposed Model}
\vspace{-10pt}
We first describe the basics of the proposed method (see Appendix~\ref{appendix:flow}). The architecture is equivalent to the standard AE formulation where it encodes the input spectra $\mathbf{x}$ to a hidden latent representation $\mathbf{\hat y} = Enc(\mathbf{x}; \theta^e)$ and then decodes to recompute the reconstructed version of the original spectra $\mathbf{\hat x} = Dec(\mathbf{\hat y}; \theta^d)$ by optimizing the learnable parameters $\theta^e$ and $\theta^d$ with some loss function.

\vspace{-3pt}
\subsection{Encoder Part}
\vspace{-7pt}
The dimensionality of data is one of the drawbacks, especially for unsupervised learning. The main reason is that a high-dimensional space causes irregular distributions of data in the learning process. Inevitably, overfitting occurs for the parameters. For additional background, we recommend~\cite{norouzi2011minimal, jegou2011product}.

For this purpose, high-dimensional input data $\mathbf{x}$ is mapped to a low-dimensional space $\mathbf{z} \in  \mathbb{R}^{M}$ where $M$ denotes the dimension of latent representations. Hence, a latent representation is computed with a set of learnable parameters $\theta^c$ as $\mathbf{z}=H(\mathbf{x};\theta^c)$. Observe that $M$ value should be selected as $M>>K$ to extract sparse representations. In particular, 1D convolution layers are utilized with batch normalization and sparse activation layers in this function. Since fewer parameters are optimized, the stack of convolutions boosts the representation capacity with deep hierarchies. Details can found in Appendix~\ref{appendix:NN} for the NN structure with input/output relations.

 Later, fractions $\mathbf{\hat{y}}$ are estimated by using these latent representations $\mathbf{z}$ with a multinomial function $\mathbf{\hat{y}}=G(\mathbf{z}; \theta^p)$ where $\theta^p$ is the learnable parameters. A NN multinomial mixture model is utilized for this case. As explained, distribution-based methods reproduce materials as they are sampled from probability distributions. Initial methods employ a unimodal distribution as:
\begin{equation}
p(\mathbf{y}|\mathbf{z}) \sim \mathcal{N}(\mathbf{z}|\mathbf{\mu},\mathbf{\Sigma})
\label{eq:unimodal}
\end{equation}

Fractions $\mathbf{y}$ are estimated from $\mathbf{z}$ by using a Gaussian distribution where $\mathbf{\mu}$ and $\mathbf{\Sigma}$ indicate the mean and covariance matrix, respectively. In fact, this formulation underestimates the solution by fitting a unimodal distribution for each material. Therefore, Eq.~\ref{eq:unimodal} is restored  with the Gaussian Mixture  Model  (GMM)  by  computing fractions with the mixture of Gaussian distributions:
\begin{equation}
p(\mathbf{y}|\mathbf{z}) \sim G(\mathbf{z}; \theta^p) = \sum_{n=1}^{N} \alpha_{n} \mathcal{N}(\mathbf{z}|\mathbf{\mu}_{n},\mathbf{\Sigma}_{n})
\label{eq:gmm2}
\end{equation}

Here, $N$ is the total number of mixture components by constraining $\alpha_{n} \geq 0$ and $\sum_{n=1}^{N}\alpha_{n}=1$. Multinomial distributions are adopted in our method because fractions can be overly represented so that spectral uncertainty is adequately handled. Furthermore, $N$ should be selected as $N>>K$ for effective representations. To this end, we mimic this approach with a set of NN models. We prefer to use NN models since the overall framework can be trained in an end-to-end manner. Note that similar attempts are also made in the literature~\cite{jalali2019efficient}.

If we extend Eq.~\ref{eq:gmm2} to separate the term into two learnable parts, this equation is expressed as $G(\mathbf{z}; \theta^p) = \sum_{n=1}^{N} \beta_{n}g_{n}(\mathbf{z})$ where $g_n(\mathbf{z})=\exp{(\frac{-1}{2}(\mathbf{z}-\mathbf{\mu}_n)^T\mathbf{\Sigma}_n^{-1}(\mathbf{z}-\mathbf{\mu}_n))}$ and $\beta_n=\frac{\alpha_{n}}{(2\pi)^\frac{N}{2}|\mathbf{\Sigma}_n|^\frac{1}{2}}$. As noticed, at least two sets of parameters must be optimized, thus two separate single-layer subnetworks are employed as $v_n(\mathbf{z}; \theta^{p1})=g_n(\mathbf{z})$ and $c_n(\mathbf{z}; \theta^{p2})=\beta_n$.

To model these terms, the quadratic form of $g_n(\mathbf{z})$ is renounced in our formation. Precisely, multivariate Mahalanobis distance is normalized with a sigmoid function to calculate the probabilistic relations. To this end, the normal distribution resembles a cumulative distribution function. Similarly, $\beta_n$ term is calculated with a NN model, and the sum-to-one constraint is retained with softmax activation. In the end, all subparts of the encoder can be trained with differentiable layers. 

\vspace{-3pt}
\subsubsection{Parameter Updates for Multinomial Mixture Kernel}
\vspace{-7pt}
Expectation-Maximization (EM) is the most popular technique to estimate GMM parameters. This approach guarantees a stable solution by optimizing the negative log-likelihood function. Note that our loss function exploits the negative log-likelihood term for parameter updates. However, the studies show that EM approach suffers to achieve a true solution for several reasons, such as the random initialization of parameters and discontinuity. For further details, we recommend to read~\cite{kolouri2017sliced, jin2016local}. 

To overcome these limitations, WGAN is employed, which provides several theoretical benefits over the classical approaches~\cite{kolouri2017sliced}. Specifically, Wasserstein distance is continuous and differentiable almost everywhere.  Also, since it has a smooth loss function, the initialization is not critical. Formally, Wasserstein distance $W(p,q)$ measures the distance between two distributions $p$ and $q$ where they generate variables as $\mathbf{x}^p$ and $\mathbf{x}^q$. Kantorovich-Rubinstein duality minimizes the distance as: 
\begin{equation}
\substack{\\ max \\ \| f \|_L \leq 1} \big( \mathbb{E}_{\mathbf{x}^p}[f({\mathbf{x}^p})] - \mathbb{E}_{\mathbf{x}^q}[f({\mathbf{x}^q})] \big)
\label{eq:wgan1}
\end{equation}

where $f(.)$ is a 1-Lipschitz function. Inspired by the structure of Variational AutoEncoder (VAE)~\cite{kingma2013auto}, this duality can be adopted to our formulation in Eqs.~\ref{eq:sulinear} and~\ref{eq:gmm2} as:
\begin{equation}
\substack{\\ max \\ \| f \|_L \leq 1} \bigg( \mathbb{E}_{\mathbf{x}}[f({\mathbf{\mathbf{x}}})] - \mathbb{E}_{\mathbf{\hat{x}} }\bigg[f(\sum_{k=1}^{K} (\mathbf{e}_k +\mathbf{\gamma}_k)\sum_{n=1}^{N} \beta_{n} g_n(\mathbf{z}))\bigg] \bigg)
\label{eq:wgan2}
\end{equation}

Here $\gamma_k$ is omitted for the simplicity and $\beta_{n}$ value is accepted as not random. Note that this simplification is resolved in the decoder part. In addition, the generated sample from multinomial components is denoted as $\mathbf{\hat{x}}$. With these observations, the formulation is simplified as:
\begin{equation}
\substack{\\ max \\ \|f \|_L \leq 1} \bigg( \mathbb{E}_{\mathbf{x}}[f({\mathbf{x}})] - \sum_{k=1}^{K} \mathbf{e}_k\sum_{n=1}^{N} \beta_n \mathbb{E}_{\mathbf{\hat{x}}}[f(g_n(\mathbf{z}))] \bigg)
\label{eq:wgan3}
\end{equation}

To this end, this notion mathematically proves that WGAN loss can be used for the parameter updates of $g_n(z)$ terms. Furthermore, an additional gradient penalty term is used for stable solutions~\cite{gulrajani2017improved}:
\begin{equation}
\mathcal{L}_{disc} = \mathbb{E}[d(\mathbf{x}; \theta^{d})] - \mathbb{E}[d(\mathbf{\hat{x}};  \theta^{d})] \hspace{1mm} \\
+ \lambda_{pq} \mathbb{E}[(\| d(\mathbf{\tilde x};  \theta^{d}) \|_2-1)^2].
\label{eq:wgan4}
\end{equation}

Here, $d(.,.)$ is the discriminative module and it is a 1-Lipschitz function. Furthermore, $\mathbf{\tilde{x}}$ is uniformly sampled data from real $\mathbf{x}$ and reconstructed $\mathbf{\hat{x}}$ data. $\lambda_{pq}$ scales the contribution of penalty term that is set to $10$. The details for the NN structure of discriminative module can be found in Appendix~\ref{appendix:NN}.

\vspace{-3pt}
\subsection{Decoder Part}
\vspace{-7pt}
As stated, in Eq.~\ref{eq:sulinear}, the formulation defines $\gamma_k$ value for spectral uncertainty. However, we omit this term in Eq.~\ref{eq:wgan3} for the simplicity. Here, we remodel this term with extra NN layers.

For this purpose, we add an uncertainty NN model $u(\mathbf{\hat{y}}, \mathbf{\eta} ; \theta^u)$ that takes the output of multinomial mixture kernel and random noise as input. This model computes an additive full-size spectral signature. Observe that there is random noise $\eta$ to mimic the spectral uncertainty for different pixels in the formulation. Lastly, WGAN loss is also used for these parameters. Moreover, we utilize a refinement NN model $u(\mathbf{\hat{y}}; \theta^r)$ that allows updates and nonlinearity for the precomputed endmembers. 

The final output of the decoder part is reformulated by rescaling the contributions of refinement and uncertainty NN models with $\lambda_r=0.05$ and $\lambda_u=0.1$, respectively:
\begin{equation}
\mathbf{\hat x} = Dec(\mathbf{\hat y}; \theta_d) = \mathbf{E} \mathbf{\hat y} + \lambda_u u(\mathbf{\hat y},\eta;\theta^u) + \lambda_r u(\mathbf{\hat y}; \theta^r)
\label{eq:decoder}
\end{equation}

\subsection{Implementation Details}
\vspace{-7pt}
For the parameter updates, spectral angle distance based loss function is utilized~\cite{ozkan2018endnet}. WGAN loss is also used for the parameters at encoder and decoder parts, as explained. Adam stochastic optimizer is used by setting $\beta_1$ as $0.7$~\cite{kingma2014adam}. 

\section{Experiments}
\vspace{-7pt}
Experiments to compare the performance of the proposed method (DSCN++) with the baseline techniques are conducted on two real datasets (Urban~\cite{zhu2014spectral} and Jasper~\cite{zhu2014structured} datasets) (see Appendix~\ref{appendix:sync} for additional experiments). For this purpose, Linear Mixture Model (LMM)~\cite{heylen2016multilinear}, Generalized Bilinear Model (GBM)~\cite{halimi2011nonlinear}, Post-Nonlinear Mixing Model (PPNM)~\cite{altmann2012supervised}, Multilinear Mixing Model (MLM)~\cite{heylen2016multilinear} and Normal Compositional Model (NCM)~\cite{stein2003application} are selected that are extensively used nonlinear and distribution-based methods in the literature. Scores of the baselines are reproduced.

Furthermore, Spatial Compositional Model (SCM)~\cite{zhou2016spatial}, Distance-MaxD (DMaxD)~\cite{heylen2011non} and Sparse AutoEncoder Network (EndNet)~\cite{ozkan2018endnet} are exploited to estimate endmember spectra from the scenes. Scores are reported in terms of Root Mean Square Error (RMSE) and tests are repeated $20$ times.

First, we test all individual steps and parameter configurations for different datasets. For clarity, EU stands for NN models that correspond to uncertainty and refinement models in the decoder part. Furthermore, WGAN indicates the use of Wasserstein GAN in parameter optimization. $N$ denotes the total number of mixture components. Experimental results are reported in Tab.~\ref{tab:conf}. An important observation is that all these contributions ultimately improve performance. In particular, both WGAN and EU steps improve the performance of DmaxD methods for both datasets. A similar observation can be made for SCM on the Urban dataset and Endnet on the Jasper dataset. Lastly, the selection of $N$ value larger than the material number $K$ ultimately advances the unmixing performance.

Second, we also compare our method with the baselines. Results are demonstrated in Tab.~\ref{tab:base}. Observe that no single baseline method can continuously yield accurate results for different scenes, even if these baselines cover both nonlinear and distribution-based methods. This observation shows the superiority of our method that can capture both nonlinearity and uncertainty conditions perfectly. Of course, our method obtains state-of-the-art performance compared to baselines by a large margin.

\begin{table}[t]
\begin{center}

\caption{RMSE performance for Urban and Jasper datasets. Best results are bold.}
\vspace{-10pt}
\begin{tabular}{c*{8}{c}}
\hline \hline
& \multicolumn{3}{c}{Urban Dataset} &  & \multicolumn{3}{c}{Jasper Dataset}\\
\hline
$\times 10^{-2}$ & DMaxD & SCM & EndNet & & DMaxD & SCM & EndNet \\
\hline

N$=4$ & 15.96 $\pm$2.6 & 13.26 $\pm$0.6 & 8.27 $\pm$0.4 & & \textbf{12.03} $\pm$0.7 & 18.10 $\pm$0.9 & 6.38 $\pm$0.7 \\
N$=8$ & \textbf{15.47} $\pm$1.9 & 13.29 $\pm$0.8 & 8.05 $\pm$0.5 & & 12.29 $\pm$0.9 & 18.33 $\pm$1.1 & 6.61 $\pm$0.7 \\
N$=16$ & 16.32 $\pm$3.2 & 13.24 $\pm$0.5 & 8.16 $\pm$0.4 & & 12.12 $\pm$0.7 & \textbf{17.92} $\pm$0.8 & \textbf{6.31} $\pm$0.7 \\
N$=24$ & 16.56 $\pm$2.1 & \textbf{13.23} $\pm$0.8 & \textbf{8.04} $\pm$0.3 & & 12.27 $\pm$0.6 & 18.03 $\pm$1.1 & 6.62 $\pm$1.1 \\
\hline \hline

w$\setminus$o EU &  34.14 $\pm$0.3  & 15.91 $\pm$0.3    & 8.38 $\pm$0.4  & &  14.78  $\pm$0.4  & 20.26 $\pm$0.4 & 8.36 $\pm$0.5  \\
w$\setminus$o WGAN       &  31.43 $\pm$1.6   & 15.76 $\pm$1.0  & 8.52 $\pm$0.5 &  &  15.63 $\pm$0.3   & 19.03 $\pm$0.4 & 10.15 $\pm$0.5 \\
\hline \hline
\end{tabular}
\label{tab:conf}

\end{center}
\end{table}

\vspace{-6pt}
\begin{table}
\begin{center}

\caption{RMSE performance on Urban and Jasper datasets. Best results are bold. }
\vspace{-6pt}
\begin{tabular}{c*{8}{c}}
\hline \hline
& \multicolumn{3}{c}{Urban Dataset} &  & \multicolumn{3}{c}{Jasper Dataset}\\
\hline
$\times 10^{-2}$ & DMaxD & SCM & EndNet & & DMaxD & SCM & EndNet \\
\hline

LMM & 31.04 & 28.00 & 27.59 & & 15.31 & 19.78 & 22.60 \\
GBM & 31.04 & 27.00 & 27.08 & & 15.98 & 19.95 & 21.91\\
PPNM & 36.54 & 28.17 & 15.06 & & 12.99 & 19.78 & 14.72 \\
MLM &  31.43 & 16.34 & 9.48 & & 16.59 & 20.73 & 8.13 \\
NCM & 31.04 &  28.00 & 27.51 & & 15.29 &  19.56 & 22.42 \\
\hline
DSCN++ & \textbf{15.47} & \textbf{13.23} & \textbf{8.04} & & \textbf{12.03} & \textbf{17.92} & \textbf{6.31} \\

\hline \hline
\end{tabular}
\label{tab:base}

\end{center}
\end{table}

\section{Conclusion}
\vspace{-7pt}
In this paper, we introduce a novel neural network (NN) framework for spectral unmixing. It is composed of 1D convolutions by addressing spectral uncertainty, which is one of the severe physical conditions frequently observed in real data. For this purpose, a novel NN layer is proposed to estimate the fractions per-pixel by leveraging the multinomial mixture kernel. Wasserstein GAN is exploited for the parameter optimization, which leads to more accurate and stable performance than other loss functions. The experimental results validate that the proposed method outperforms baselines by a large margin. 

\bibliographystyle{plain}
\bibliography{neurips_2020} 

\normalsize
\newpage 
\appendix

\section{Appendix For Spectral Uncertainty}
\label{appendix:su}

\begin{figure}[h]
\centering

\includegraphics{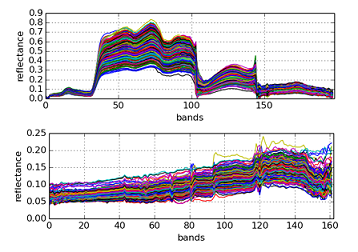}
\caption{Spectral signatures from Jasper (first row) and Urban (second row) data correspond to two different pure materials.}

\label{fig:svar}
\end{figure}

As explained, spectral images can be significantly affected by variations in atmospheric, illumination, or environmental conditions. Hence, spectral uncertainty ultimately influences the mixture of materials, especially for real data. Fig.~\ref{fig:svar} illustrates the pure material samples from real datasets. Observe that this uncertainty can be tedious for spectral data. 

\section{Appendix For Model Flow}
\label{appendix:flow}

\begin{figure}[h]
\centering

\includegraphics{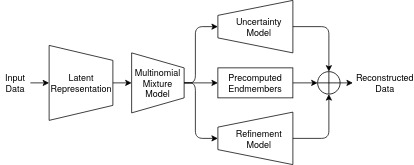}
\caption{Flow of the proposed model.}

\label{fig:flow}
\end{figure}

The overall flow of the proposed method is illustrated in Fig.~\ref{fig:flow}. Our model takes spectral data $\mathbf{x}$ as input and computes a latent representation $\mathbf{z}$. Later, this representation is modeled with a multinomial mixture model to estimate fractions $\mathbf{y}$.  These fractions are fed to two NN models with the matrix multiplication of precomputed endmembers in the decoder part.  In the end,  individual outputs are summed, and the input is reconstructed. 

\section{Appendix For NN Structures}
\label{appendix:NN}

In our method, two deep NN models are primarily utilized. For the encoder part, a stack of 1D convolution layers is used to transform high-dimensional spectral data to a low-dimensional space, as illustrated in Tab.~\ref{tab:DSCN}. Observe that the order of convolution and normalization layers is modified. This modification enhances the selectivity of representations by feeding more statistical knowledge to the convolution layers. Furthermore, Inception module that has multiple kernels of different sizes is also integrated. Hence, a rich representation from multiple kernel sizes can be computed. 

\begin{table} [h]
\begin{center}

\caption{The NN Structure of the encoder part. The notation of Conv1D(W,K,S) indicates 1D convolution whose output dimension, kernel size and stride are W, K and S, respectively. Furthermore, Nrm and PReLU denote normalization and activation layers. AvgP(K) corresponds to the average pooling layer with kernel size and stride K.}

\begin{tabular}{c*{4}{c}}
\hline \hline
\textbf{Index} & \textbf{Index of Inputs} & \textbf{Operation(s)} & \textbf{Output Shape} \\
\hline
(1) & - & Input Spectra &  D \\
(2) & (1) & Conv1D(10, 21, 1) & D $\times$ 10 \\
(3) & (2) & PReLU + AvgP(5) + Nrm & D/5 $\times$ 10 \\
(4) & (3) & Conv1D(10, 3, 1) & D/5 $\times$ 10 \\
(5) & (3) & Conv1D(10, 5, 1) & D/5 $\times$ 10 \\
(6) & (3) & Conv1D(10, 7, 1) & D/5 $\times$ 10 \\
(7) & (4),(5),(6) & Concat & D/5 $\times$ 30 \\
(8) & (7) & PReLU + AvgP(2) + Nrm & D/10 $\times$ 30 \\
(9) & (8) & Conv1D(5, 3, 1) & D/10 $\times$ 10 \\
(10) & (9) & PReLU + AvgP(2) + Nrm & D/20 $\times$ 10 \\
(11) & (10) & Linear(M) + LReLU & M \\
\hline \hline
\end{tabular}
\label{tab:DSCN}

\end{center}
\end{table}

\begin{figure}[h]
\centering

\includegraphics{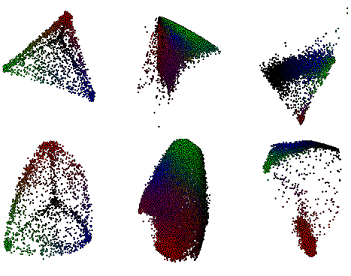}
\caption{Data distributions before (first raw) and after (second raw) applying the encoder part.}

\label{fig:hidden}
\end{figure}

To demonstrate the superiority of this model, Fig.~\ref{fig:hidden} visualizes the distributions of input data and latent representation transformed by our method. Note that colors are discriminative based on their real fractions. Our method can blindly achieve more separable space for spectral data.

Moreover, we utilize a deep NN model for the discriminative part of WGAN. This structure is also given in Tab.~\ref{tab:disc}. In particular, PatchGAN is adopted in our method. Hence, local patches are fed to the discriminative model than full-dimensional data. Local characteristics (i.e., local spectral uncertainty) are practically captured.

\begin{table}
\begin{center}

\caption{NN Structure of discriminative network.}

\begin{tabular}{c*{4}{c}}
\hline \hline
\textbf{Index} & \textbf{Index of Inputs} & \textbf{Operation(s)} & \textbf{Output Shape} \\
\hline
(1) & - & Input Spectra &  D \\
(2) & (1) & Conv1D(5, 21, 5) & D/5 $\times$ 5 \\
(3) & (2) & Nrm + PReLU & D/5 $\times$ 5 \\
(4) & (3) & Conv1D(10, 5, 2) & D/10 $\times$ 10 \\
(5) & (4) & Nrm + PReLU & D/10 $\times$ 10 \\
(6) & (5) & Conv1D(20, 5, 2) & D/20 $\times$ 20 \\
(7) & (6) & Nrm + PReLU & D/20 $\times$ 20 \\
(8) & (7) & Linear(5) & D/20 $\times$ 5 \\
\hline \hline
\end{tabular}
\label{tab:disc}

\end{center}
\end{table}

\section{Appendix For Additional Experiments on Synthetic Data}
\label{appendix:sync}

Our method is also tested on a synthetic dataset released for the experiments in~\cite{zhou2018gaussian}. The dataset consists of four constituent materials from the ASTER spectral library: limestone, basalt, concrete and asphalt. For the scene, limestone is set as the background material; the rest is randomly placed to the image by maintaining the Gaussian-shaped distributions.

\begin{figure}[h]
\centering
\includegraphics{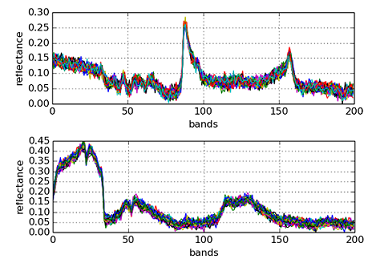}
\caption{Spectra of pixels from the Synthetic dataset.}
\label{fig:noise_synt}
\end{figure}

Even if synthetic data is conventional to evaluate the methods, it is not suitable to have all adverse cases exhibited on real data. Hence, it is hard to simulate nonlinearity and spectral uncertainty on the data. In particular, additive noise is inserted to simulate spectral uncertainty, which cannot reflect the actual physical conditions. Fig.~\ref{fig:noise_synt} illustrates samples from two pure materials on this synthetic dataset. Unlike real data, these samples lead to corrupted signatures.

Experimental results are reported in Tab.~\ref{tab:synt_conf}. If we do not use EU and WGAN steps, the model achieves the best performance. This result makes sense since spectral uncertainty is simulated as random additive noise. Hence, WGAN loss eventually learns this noise characteristics instead of the real spectral uncertainty. This characteristic drastically reduces the performance and undermines the model. However, as explained, the actual capacity of these steps on real data is obvious.

We also compare our method with baselines. In Tab.~\ref{tab:synt_base}, these results show that the proposed method performs better even if the full capacity of model is not used.

\begin{table}[h]
\begin{center}

\caption{RMSE performance on synthetic dataset. Best results are bold.}

\begin{tabular}{c*{5}{c}}
\hline \hline
$\times 10^{-2}$ & DMaxD & SCM & EndNet \\
\hline

N$=4$   & 8.32 $\pm$1.4 & 4.47 $\pm$0.7 & 9.20 $\pm$1.4 \\
N$=8$   & 8.15 $\pm$1.4 & 4.36 $\pm$0.9 & 9.14 $\pm$1.0 \\
N$=16$ & 8.18 $\pm$1.0 & 3.97 $\pm$0.5 & 9.06 $\pm$1.1 \\
N$=24$ & 7.55 $\pm$1.1 & 4.42 $\pm$0.6 & 9.51 $\pm$1.2 \\
\hline
\hline
w$\setminus$o EU       & 5.43  $\pm$0.3  & 3.81 $\pm$0.9  & \textbf{6.43} $\pm$0.4 \\
w$\setminus$o WGAN & \textbf{5.23}  $ \pm$0.4 & \textbf{3.22} $\pm$1.2  & 8.88 $\pm$0.3  \\

\hline \hline
\end{tabular}
\label{tab:synt_conf}

\end{center}
\end{table}

\begin{table}[h]
\begin{center}

\caption{RMSE performance on synthetic dataset. Best results are bold.}

\begin{tabular}{c*{4}{c}}
\hline \hline
$\times 10^{-2}$ & DMaxD & SCM & EndNet \\
\hline

LMM & 5.17 & 3.36 & 17.62 \\
GBM & 5.62 & 3.26 & 17.39 \\
PPNM & 7.66 & 4.01 & 9.34\\
MLM & 6.52 & 3.85 & 6.71 \\
NCM & \textbf{4.09} & 3.85 & 11.17 \\
\hline
DSCN++ & 5.23 & \textbf{3.22} & \textbf{6.43}  \\

\hline \hline
\end{tabular}
\label{tab:synt_base}

\end{center}
\end{table}

\end{document}